\documentclass{article}

\usepackage{arxiv}

\usepackage[utf8]{inputenc} 
\usepackage[T1]{fontenc}    
\usepackage{hyperref}       
\usepackage{url}            
\usepackage{booktabs}       
\usepackage{amsfonts}       
\usepackage{nicefrac}       
\usepackage{microtype}      
\usepackage{lipsum}
\usepackage{graphicx}

\usepackage{algorithm}
\usepackage{algpseudocode}
\usepackage{caption}
\usepackage{subcaption}
\usepackage{amsmath}
\usepackage{hyperref}

\graphicspath{ {./images/} }

\title{Edge AI Collaborative Learning:\\Bayesian Approaches to Uncertainty Estimation}

\author{
 Gleb Radchenko \\
  Silicon Austria Labs\\
  Sandgasse 34, 8010 Graz, Austria\\
  \texttt{gleb.radchenko@silicon-austria.com} \\
   \And
 Victoria Andrea Fill \\
  FH Joanneum\\
   Alte Poststraße 149, 8020 Graz, Austria\\
  \texttt{victoria.fill@edu.fh-joanneum.at} \\
}

\begin{document}
\maketitle
\begin{abstract}
Recent advancements in edge computing have significantly enhanced the AI capabilities of Internet of Things (IoT) devices. However, these advancements introduce new challenges in knowledge exchange and resource management, particularly addressing the spatiotemporal data locality in edge computing environments. This study examines algorithms and methods for deploying distributed machine learning within autonomous, network-capable, AI-enabled edge devices. We focus on determining confidence levels in learning outcomes considering the spatial variability of data encountered by independent agents. Using collaborative mapping as a case study, we explore the application of the Distributed Neural Network Optimization (DiNNO) algorithm extended with Bayesian neural networks (BNNs) for uncertainty estimation. We implement a 3D environment simulation using the Webots platform to simulate collaborative mapping tasks, decouple the DiNNO algorithm into independent processes for asynchronous network communication in distributed learning, and integrate distributed uncertainty estimation using BNNs. Our experiments demonstrate that BNNs can effectively support uncertainty estimation in a distributed learning context, with precise tuning of learning hyperparameters crucial for effective uncertainty assessment. Notably, applying Kullback–Leibler divergence for parameter regularization resulted in a 12-30\% reduction in validation loss during distributed BNN training compared to other regularization strategies. 
\end{abstract}


\section{Introduction}
Recent advancements in edge computing have significantly increased the computational capabilities of Internet of Things (IoT) devices, enabling more complex data processing at the network edge \cite{Liang2023,Parmar2023,Wang2022}. This evolution in edge device capabilities introduces new challenges and opportunities in distributed data processing and analysis.

Integrating advanced processing capabilities into edge devices aims to optimize local data handling and enhance device autonomy. This approach is particularly relevant when low-latency processing, data privacy, and network efficiency are critical. However, implementing machine learning (ML) algorithms on edge devices introduces several challenges:
\begin{itemize}
    \item \textbf{Distributed Learning:} How can we effectively implement distributed learning on edge devices, transitioning from traditional inference-only models to active participation in the learning process while ensuring data privacy and efficient knowledge sharing?~\cite{Lim2020}
    \item \textbf{Resource Management and Communication Efficiency:} How do we optimally manage limited resources (e.g., power, processing capabilities, network bandwidth) on edge devices while maintaining efficient communication for data and model updates in dynamic network conditions?~\cite{Samie2016,Park2021}
    \item \textbf{Spatio-Temporal Locality and Non-IID Data:} What strategies most effectively incorporate localized and non-IID data to improve accuracy in distributed ML models? Evaluating the uncertainty at the level of both the individual agent and the aggregated model is essential.~\cite{Jospin2022,Yang2023}
\end{itemize}

\subsection{Contribution}
This study focuses on algorithms and methods for deploying distributed ML within autonomous, network-capable, sensor-equipped, AI-enabled edge devices. Specifically, it addresses determining confidence levels in learning outcomes, considering the spatial variability of data sets encountered by independent agents. To address this issue, we investigate the potential of the Distributed Neural Network Optimization (DiNNO) algorithm~\cite{Yu2022}, aiming to extend it for organizing distributed data processing and uncertainty estimation using Bayesian neural networks (BNN)~\cite{Jospin2022}.

Our research examines the interactions of AI-enabled edge devices within a collaborative mapping task. To meet our objectives, we need to address the following tasks:
\begin{enumerate}
    \item Decoupling the DiNNO algorithm implementation into independent processes, enabling asynchronous network communication for distributed learning.
    \item Integrating distributed uncertainty estimation into the resulting neural network models by applying BNNs and evaluating the applicability of BNNs in the context of distributed learning.
    \item Implementing the proposed approaches within a case study: simulation of robots, augmented with LiDAR sensors for environmental mapping, navigating a 3D environment using the Webots platform.
    \item Evaluating the effectiveness of distributed uncertainty estimation using BNNs in a distributed learning context, including tuning of learning hyperparameters and applying Kullback–Leibler divergence for NN parameter regularization.
\end{enumerate}

This paper is an extended version of the work entitled "Uncertainty Estimation in Multi-Agent Distributed Learning for AI-Enabled Edge Devices"~\cite{Radchenko2024}, which was presented at the 14\textsuperscript{th} International Conference on Cloud Computing and Services Science (CLOSER). This extended version expands upon the previous work by providing an expanded analysis of distributed ML algorithms and evaluating the feasibility of an online learning approach within the collaborative mapping case.

\subsection{Outline}
The remainder of this paper is structured as follows. In Section 2, we provide an overview of the related work in distributed edge-based ML methods and Bayesian Neural Networks. Section 3 presents the collaborative mapping case study. Sections 4 and 5 elaborate on the distributed edge learning approach and uncertainty estimation method. In Section 6, we describe the implementation and evaluation of the proposed methods. Finally, Section 7 presents conclusions and suggestions for future work.

\section{Related Work}
\subsection{Edge Learning Methods}

Edge computing has recently enabled local devices to perform AI tasks that were once limited to centralized systems. Initially, these devices only supported edge-based inference, using pre-trained models to process local data~\cite{Li2020,Shao2020}. However, with the development of Edge Learning (EL), these devices can now both train and update models locally~\cite{Merenda2020}. By enabling data processing directly where it is generated, EL improves adaptability and personalization. Moreover, because the data remains on edge devices and is not transferred to centralized cloud servers, EL reduces bandwidth usage and mitigates privacy risks~\cite{Tak2021,Zhang2022,Sudharsan2020}. This advancement not only enhances the capabilities of edge devices but also marks a significant step forward in the field of distributed ML~\cite{Jiang2022}.

Distributed ML algorithms can be categorized based on their communication mechanisms, which primarily involve the exchange of model parameters, model outputs, or hidden activations. These exchanges are typically facilitated through peer-to-peer or client-server architectures~\cite{Park2021}. Among the various approaches to organizing distributed and edge learning, the most widely used methods include Federated Learning, Federated Distillation, Split Learning, and a range of decentralized learning techniques based on the Alternating Direction Method of Multipliers and its derivatives.

\subsubsection{Federated Learning (FL)} 
\begin{figure}[htbp]
    \centering
    \begin{subfigure}[b]{0.8\textwidth}
        \includegraphics[width=\textwidth]{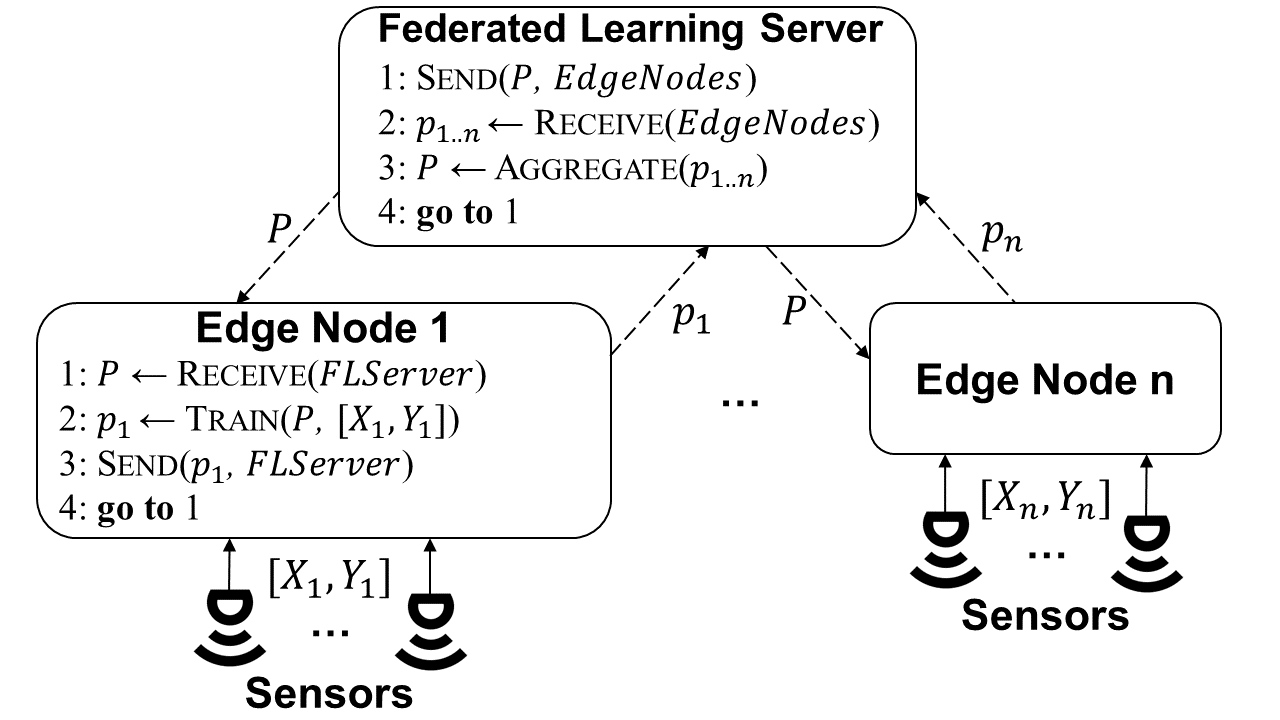}
        \caption{Federated Learning}
        \label{subfig:fl}
    \end{subfigure}
    \begin{subfigure}[b]{0.8\textwidth}
        \includegraphics[width=\textwidth]{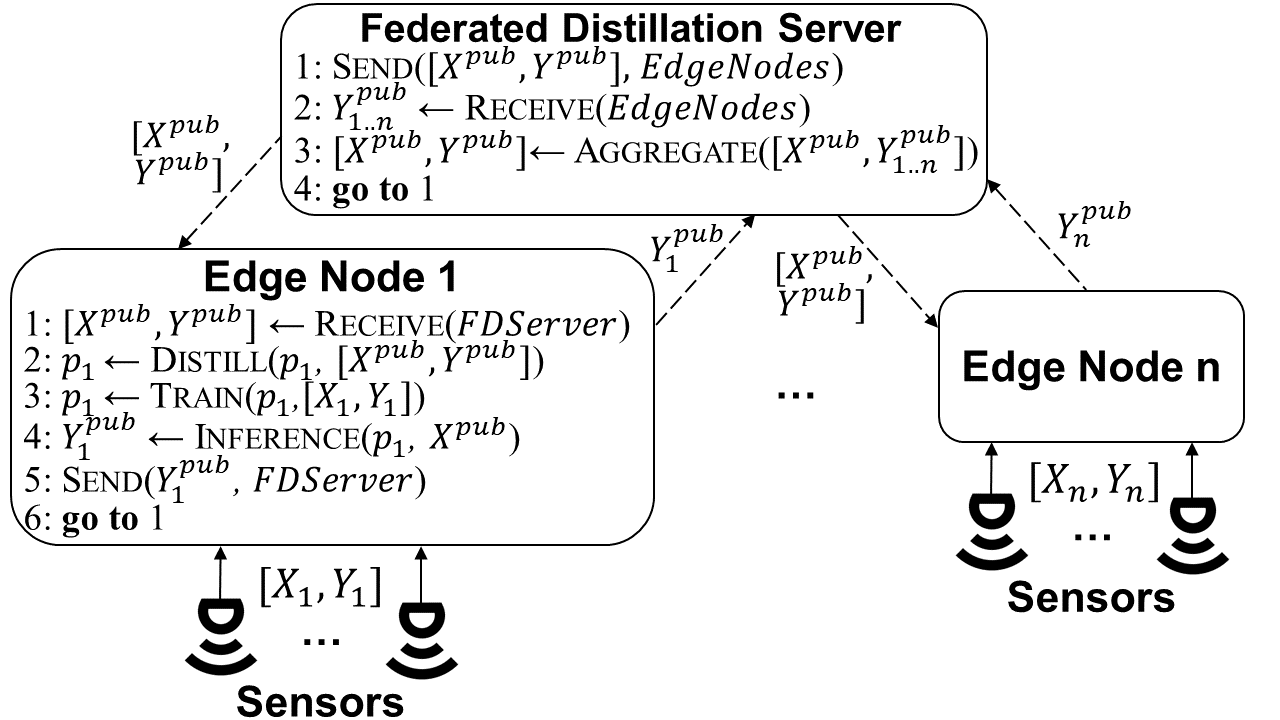}
        \caption{Federated Distillation}
        \label{subfig:fd}
    \end{subfigure}
    \caption{Graphical representation of (a) Federated Learning and (b) Federated Distillation process.}
    \label{fig:fl_fd}
\end{figure}

(see Fig.~\ref{subfig:fl}) orchestrates the periodic transmission of local training parameters from edge nodes to the Federated Learning Server. The server sends the global model $P$ to all edge nodes. Each edge node then trains the model on its local data $[X_i, Y_i]$ collected from sensors, updating its local parameters $p_i$. These local parameters are returned to the server and aggregated (\textsc{Aggregate}$(p_{1..n})$) to update the global model $P$. This updated global model is then distributed back to the edge nodes, and the process repeats iteratively.

FL has emerged as a promising strategy for EL, offering a viable solution to various challenges in edge computing environments, including the potential to preserve data privacy by avoiding the need for raw data exchange. Originally proposed to enable distributed learning using mobile devices~\cite{Brendan_McMahan2017}, FL can enhance communication efficiency by allowing for adjustable transmission intervals, which helps to optimize resource usage and reduce bandwidth demands~\cite{Lim2020}. The authors of \cite{Abreha2022} further emphasize that FL addresses issues such as unwanted bandwidth loss, data privacy concerns, and legal compliance. They highlight that FL facilitates the co-training of models across distributed clients — from mobile phones to automobiles and hospitals — via a centralized server while ensuring that the data remains localized. The authors of \cite{Nguyen2022} propose an extension of the FL model called FedFog designed to enable FL over a wireless fog-cloud system. The authors address key challenges such as non-identically distributed data and user heterogeneity. The FedFog algorithm performs local aggregation of gradient parameters at fog servers and a global training update in the cloud.

On the other hand, recent research has highlighted several limitations of FL, particularly in terms of privacy and robustness. For instance, FL is vulnerable to malicious servers and actors, which can compromise the privacy of sensitive information~\cite{Melis2019}. Furthermore, the work of Lyu et al.~\cite{Lyu2024} discusses how these vulnerabilities might undermine the privacy guarantees typically associated with~FL.

\subsubsection{Federated Distillation (FD)} (see Fig.~\ref{subfig:fd}) takes a distinct approach to communicating knowledge obtained during local training. Unlike traditional FL, which involves transmitting model parameters, FD leverages model distillation. In this technique, knowledge is transferred from one model (the "teacher") to another (the "student") by training the student on the teacher's outputs (soft-label predictions) rather than the raw data. A widely known application of knowledge distillation is model compression, through which the knowledge of a large pre-trained model may be transferred to a smaller one~\cite{Ahn2019}.

Instead of sharing the parameterization of locally trained models, the knowledge in FD is communicated through soft-label predictions on a public distillation dataset~\cite{Sattler2022}. FD orchestrates an iterative knowledge-sharing process between a central server and multiple edge nodes. At the start of each round, the server sends aggregated soft labels $Y^{pub}$ on a public dataset $X^{pub}$ to the participating edge nodes. Each edge node then performs the following steps:

\begin{enumerate}
\item \textbf{Distillation:} The node updates its local model using the soft labels received, incorporating global knowledge.
\item \textbf{Local Training:} The node further refines the distilled model on its private local data $[X_i, Y_i]$, improving the local model’s performance.
\item \textbf{Prediction:} The node generates new soft labels $Y_i^{pub}$ by running the refined model on the public dataset $X^{pub}$.
\item \textbf{Communication:} These newly generated soft labels $Y_i^{pub}$ are returned to the server.
\end{enumerate}

The server aggregates the soft labels received from all participating nodes (\textsc{Aggregate}\( [X^{pub}, Y^{pub}_{1..n}] \)) to update the global public dataset. This iterative process improves collective knowledge while ensuring that raw local data remains private. 

The soft-label exchange in FD significantly reduces the data that needs to be transferred during training compared to the exchange of full model parameters in FL. According to the authors of~\cite{Wu2022}, FD can reduce communication costs by up to 94.89\% compared to the classical FedAvg~\cite{Brendan_McMahan2017} FL method. Another advantage of FD is its flexibility in accommodating different model architectures on edge nodes. Since model parameters are not exchanged between the edge nodes and the FD server, the models on the server and the edge nodes can vary significantly in structure. Additionally, FD effectively addresses challenges in collaborative learning under non-independent and identically distributed (non-IID) data conditions. The Selective-FD algorithm proposed in~\cite{Shao2024} demonstrates high accuracy even in severe non-IID environments, enabling a communication-efficient and heterogeneity-adaptive EL.

\subsubsection{Split Learning (SL)} 
\begin{figure}[htbp]
    \centering
    \begin{subfigure}[b]{0.8\textwidth}
        \includegraphics[width=\textwidth]{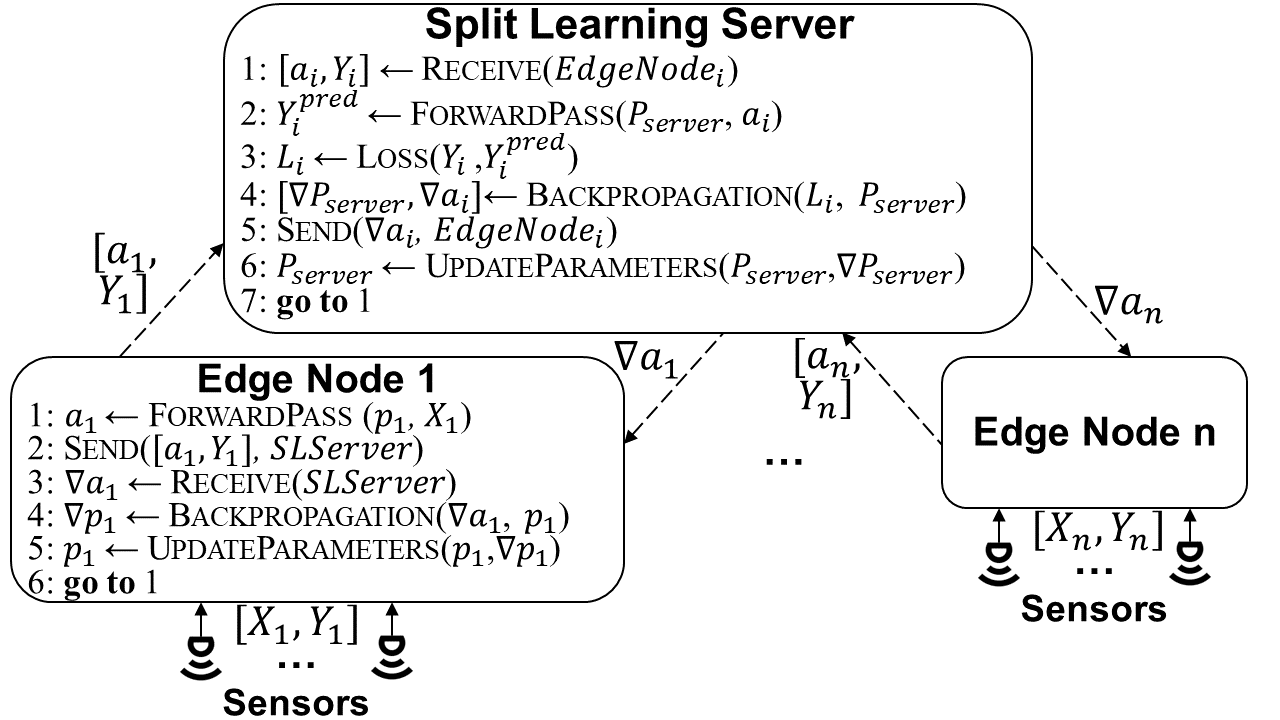}
        \caption{Split Learning}
        \label{subfig:sl}
    \end{subfigure}
    \begin{subfigure}[b]{0.8\textwidth}
        \includegraphics[width=\textwidth]{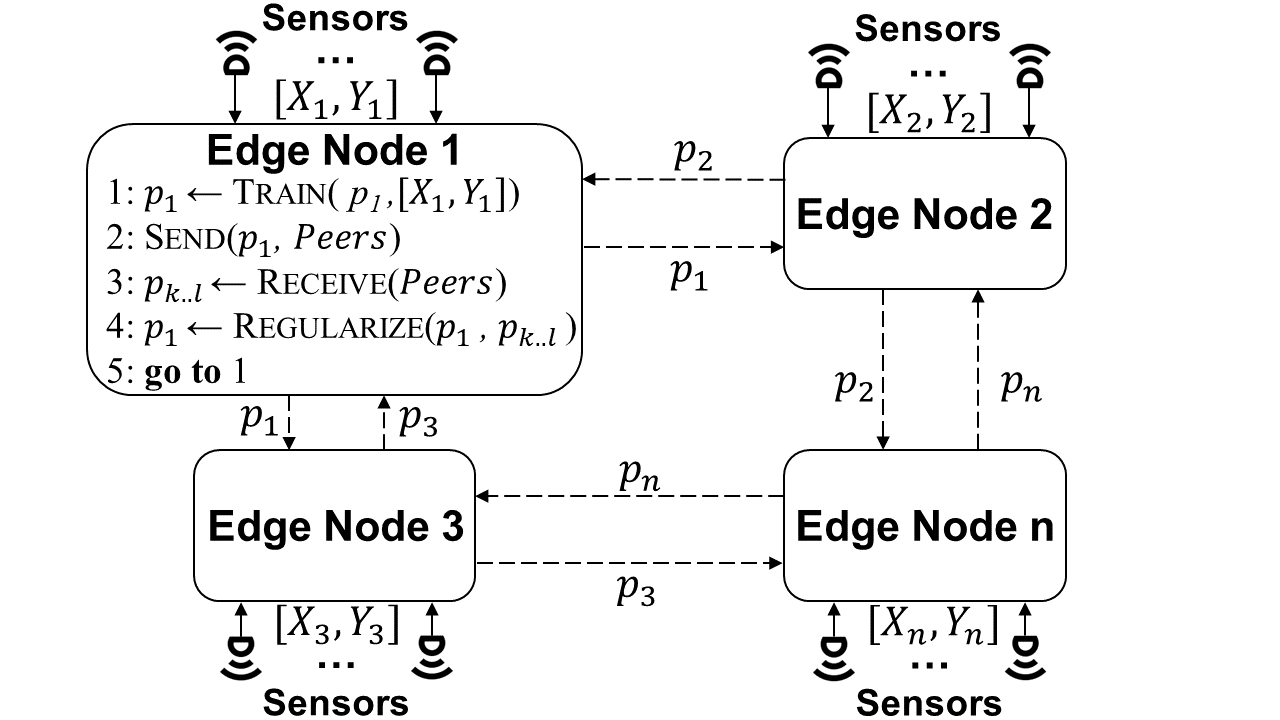}
        \caption{ADMM-based P2P learning}
        \label{subfig:ADMM}
    \end{subfigure}

    \caption{Graphical representation of (a) Split Learning and (b) a decentralized learning process using the Alternating Direction Method of Multipliers.}
    \label{fig:uncertainty_estimation}
\end{figure}

(see Fig.~\ref{subfig:sl}) partitions a multi-layer neural network (NN) into segments, enabling the training of large-scale deep NNs that exceed the memory capacities of a single edge device~\cite{Vepakomma2018}. This approach divides the NN into two segments: a lower NN segment $p_i$, which resides on the edge devices containing the raw data, and an upper NN segment $P_{server}$, hosted on a parameter server~\cite{Gupta2018}. The NN cut layer serves as the boundary between these segments.

During the forward pass, the edge devices compute the activations at the NN cut layer and transmit these activations $a_i$, along with the true labels $Y_i$, to the parameter server. The parameter server then uses these activations as inputs to the upper NN segment to continue the forward pass and compute the final predictions $Y_i^{pred}$. Subsequently, the server calculates the loss $L_i$ by comparing the predictions with the true labels and initiates the backward pass.

The gradients $ \nabla a_i $, computed at the NN cut layer during backpropagation, are returned to the edge devices. The edge devices then use these gradients to update the weights of the lower NN segments, completing the training loop. This iterative process continues until the model converges.

While this description covers the most basic form of Split Learning, it’s important to note that several SL implementations differ in how the NN is partitioned and distributed between the edge and server nodes~\cite{Ha2021,Wu2023}. These variants address different challenges and trade-offs, particularly regarding  communication efficiency, which remains an active area of research~\cite{Koda2020}.

\subsubsection{Alternating Direction Method of Multipliers (ADMM)} is a general-purpose decentralized convex optimization algorithm, particularly well suited to problems in applied statistics and machine learning. It takes the form of a de\-com\-po\-sition-coordination procedure, in which the solutions to small local subproblems are coordinated to find a solution to a large global problem~\cite{Boyd2010}. Distributed machine learning algorithms derived from ADMM, such as Distributed Neural Network Optimization (DiNNO)~\cite{Yu2022} and Group Alternating Direction Method of Multipliers (GADMM)~\cite{Elgabli2020}, enable decentralized learning without the need for a coordinating server by allowing direct communication between the edge nodes in a peer-to-peer (P2P) manner (see Fig.~\ref{subfig:ADMM}). The strong convergence properties of ADMM ensure that all learning P2P agents eventually reach a consensus on the machine-learning model parameters.

In these methods, each edge node independently trains its local model parameters $p_i$ on its private data $[X_i, Y_i]$ collected from sensors. The trained parameters $p_i$ are then shared with neighboring edge nodes. Each node receives the parameters $p_{k..l}$ from its peers, which are then used to regularize its model by incorporating the knowledge from the neighboring nodes. This process is repeated as nodes exchange and regularize their model parameters.

One of the critical issues with such P2P learning methods is the communication overhead, which is proportional to the number of model parameters and learning agents. This overhead can limit the effectiveness of ADMM-derived methods in supporting deep NNs~\cite{Elgabli2020}. However, despite this limitation, ADMM-derived methods offer a decentralized alternative to FL, particularly when a central server is unavailable or impractical.

\subsection{Uncertainty Estimation and Bayesian Neural Networks}
\subsubsection{Bayesian Neural Networks}
In a conventional NN architecture, a linear neuron is characterized by a weight ($w$), a bias ($b$), and an activation function ($f_{act}$). Given an input $x$, a single linear neuron performs the following operation:
\begin{equation}
y = f_{act}(w \cdot x + b)
\end{equation}
where $y$ is the output of the neuron. 

However, as we explore more complex and uncertain environments, the deterministic nature of classical NNs becomes a limitation. The ability of NNs to generalize to data that lies outside the training distribution remains an area of ongoing research. The potential lack of generalization, coupled with the inherent instability of NNs, can lead to the appearance of false structures in their predictions. These false structures, often referred to as hallucinations, may occur when the reconstruction method incorrectly estimates parts of the initial data set that either did not contribute to the observed measurement data or cannot be recovered in a stable manner~\cite{Bhadra2021}. A promising approach to addressing these challenges is using Bayesian neural networks, which incorporate uncertainty estimates directly into their predictions.

Bayesian Neural Networks (BNNs) adopt a Bayesian framework to train neural networks with stochastic behavior~\cite{Jospin2022}. Instead of relying on fixed, deterministic values for weights and biases, BNNs employ probability distributions, typically denoted as $P(w)$ for weights and $P(b)$ for biases. These distributions are often approximated by Gaussian distributions, with the mean and standard deviation determined from the training data. As a result, a Bayesian neuron does not produce a single output but rather a range of possible values. The operation of a Bayesian Linear neuron can thus be described as:
\begin{equation}
P(y|x) = f_{act}(P(w) \times x + P(b))
\end{equation}

In the context of BNNs, the Gaussian distributions for both weights and biases are characterized by a mean ($\mu$) and a standard deviation ($\sigma$). Specifically, the weight distribution $P(w)$ is modeled as a Gaussian with a mean $w_{\mu}$ and a standard deviation $w_{\sigma}$, where:
\begin{equation}
w_{\sigma} = \log(1+e^{w_{\rho}})
\end{equation}
The parameter $w_{\rho}$ ensures that the standard deviation remains positive. Similarly, the bias distribution $P(b)$ is represented by a Gaussian with a mean $b_{\mu}$ and a standard deviation $b_{\sigma}$, defined as:
\begin{equation}
b_{\sigma} = \log(1+e^{b_{\rho}})
\end{equation}

In the forward pass of a Bayesian neuron, weights and biases are drawn from their respective probability distributions for each neuron. These sampled values are then utilized to calculate the neuron’s output. Throughout the training process, the parameters $w_{\mu}$, $w_{\rho}$, $b_{\mu}$, and $b_{\rho}$ are adjusted to enhance the overall performance of the network.

Unlike traditional neural networks, where repeated forward passes yield the same output, BNNs generate different outputs with each forward pass due to the stochastic nature of the sampled weights and biases. After multiple passes, the mean and standard deviation of the outputs can be calculated, effectively capturing the uncertainty in the predictions. This ability to quantify uncertainty provides BNNs with the advantage of offering detailed insights into how confident the model is about each prediction.

\subsubsection{Kullback-Leibler Divergence}
Kullback-Leibler Divergence (KL Divergence) \cite{Claici2020,Kullback1951} allows us to measure the difference between the Gaussian distributions representing the parameters in a BNN. KL Divergence quantifies the dissimilarity between two probability distributions and is calculated as:
\begin{equation}
D_{KL}(g||h) = \int g(x) \log\left(\frac{g(x)}{h(x)}\right) dx
\end{equation}
where $g(x)$ and $h(x)$ are two probability density functions over the same domain. 

As explained in~\cite{Belov2011}, when the distributions $N_{0}(\mu_{0}, \sigma_{0})$ and $N_{1}(\mu_{1}, \sigma_{1})$ are both normal, equation (5) may be reduced to:
\begin{equation}
D_{KL}(N_{0}||N_{1}) = \frac{1}{2} \left[ \log \left(\frac{\sigma_{1}^{2}}{\sigma_{0}^{2}}\right) + \frac{\sigma_{0}^{2} + (\mu_{0} - \mu_{1})^{2}}{\sigma_{1}^{2}} - 1 \right]
\end{equation}

In the context of BNNs, KL Divergence is applied to quantify the deviation of the network's parameter distributions from a predefined prior distribution. The total loss in a BNN is typically expressed as:
\begin{equation}
{total}_{loss} = {base}_{loss} + {kl}_{weight} \times {kl}_{loss}
\end{equation}
where ${base}_{loss}$ corresponds to the standard loss function, such as Binary Cross-Entropy or Mean Squared Error; the term ${kl}_{weight}$ represents a hyperparameter that controls the influence of uncertainty on the model's predictions; and ${kl}_{loss}$ represents the sum of the KL Divergence between the distributions of the BNN parameters $N_{0}(\mu_{0}, \sigma_{0})$ and a specified normal distribution $N_{1}(\mu_{1}, \sigma_{1})$.

\section{Collaborative Mapping Case Study}
For our case study on edge AI collaborative learning, we focused on addressing a collaborative environment mapping problem using a network of robotic edge devices. These autonomous robots, each starting from different locations, work together to construct a comprehensive map of their surroundings by utilizing onboard sensors and sharing knowledge with each other.

\begin{figure}[htbp]
    \centering
    \includegraphics[width=0.5\textwidth]{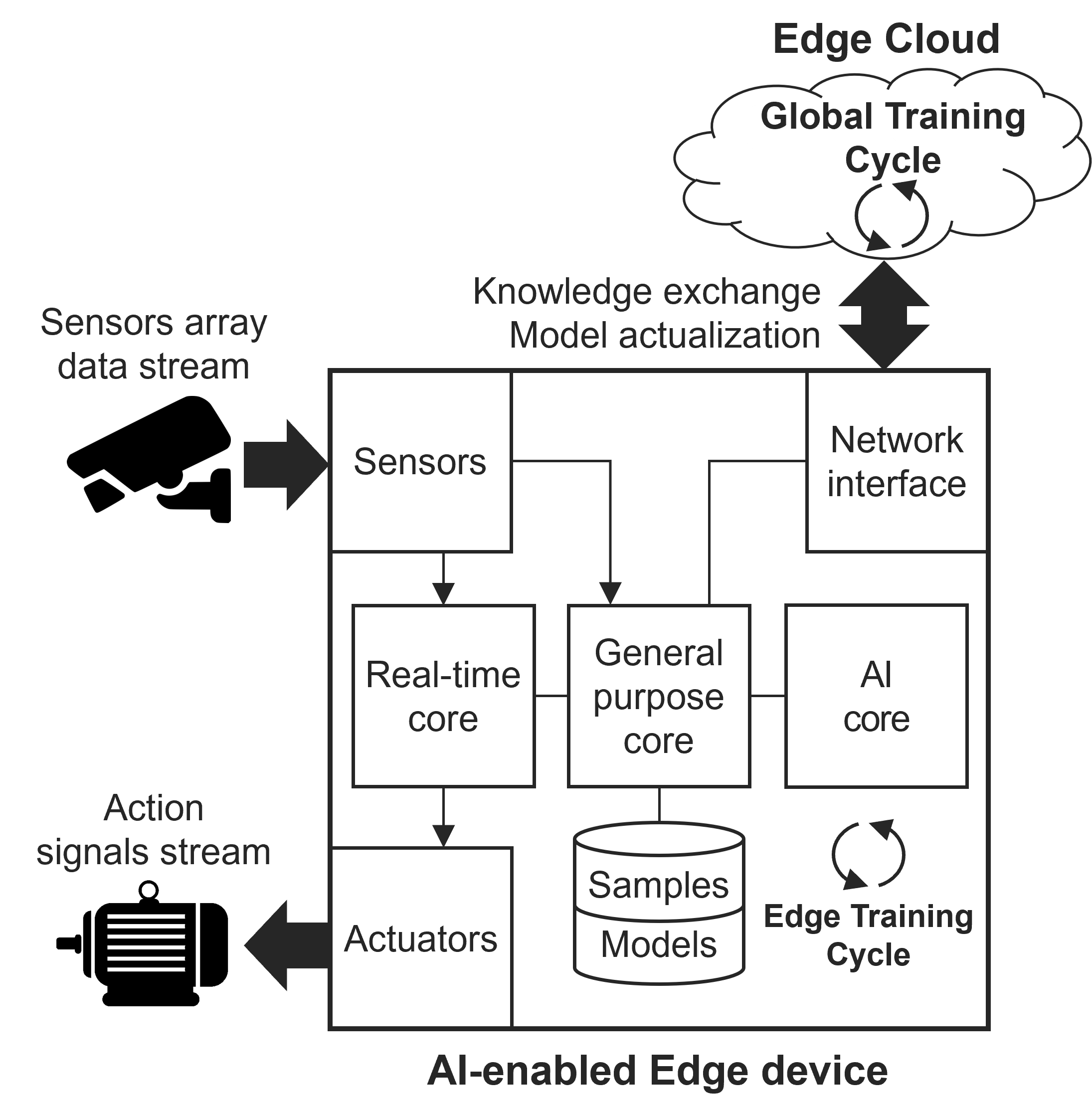}
    \caption{Components of AI-Enabled edge device~\cite{Radchenko2024}.}
    \label{fig:components}
\end{figure}

Each robot is designed to update its local ML model with new data acquired from its sensors while also communicating with other robots via a network interface. The architecture of these AI-enabled edge devices is provided in Figure~\ref{fig:components}. Each edge device is equipped with specialized computational cores tailored to handle specific tasks:
\begin{itemize}
    \item Real-time core: Handles immediate sensor data processing and controls the actuators, ensuring timely responses to environmental changes.
    \item General-purpose core: Manages overall device operations and coordinates between different components.
    \item AI core: Supports the edge training cycle by processing the data, updating the model, and facilitating knowledge exchange with other devices in the network.
\end{itemize}
These components work together to ensure that each robot can process data locally and interact with its peers to contribute to a global understanding of the environment.

To address the distributed machine learning challenge in our study, we employed the Distributed Neural Network Optimization (DiNNO) algorithm~\cite{Yu2022}. DiNNO facilitates decentralized learning by allowing each robot to optimize its NN model locally and iteratively share the learned parameters with its neighbors. Unlike centralized methods that require aggregating all data at a central node, DiNNO operates over a mesh network where robots communicate directly, ensuring robustness against node failures and preserving data privacy.

DiNNO builds on the ADMM algorithm, allowing it to efficiently converge to a consensus on the NN parameters across all robots. The algorithm's ability to work with time-varying communication graphs and streaming data is particularly suited to the dynamic and distributed nature of multi-robot systems. Each robot refines its NN model using local sensory inputs. Then, it exchanges updated parameters with its neighbors, implementing a collective learning process that eventually aligns the models across the entire network.

\begin{figure}[htbp]
    \centering
    \includegraphics[width=0.5\textwidth]{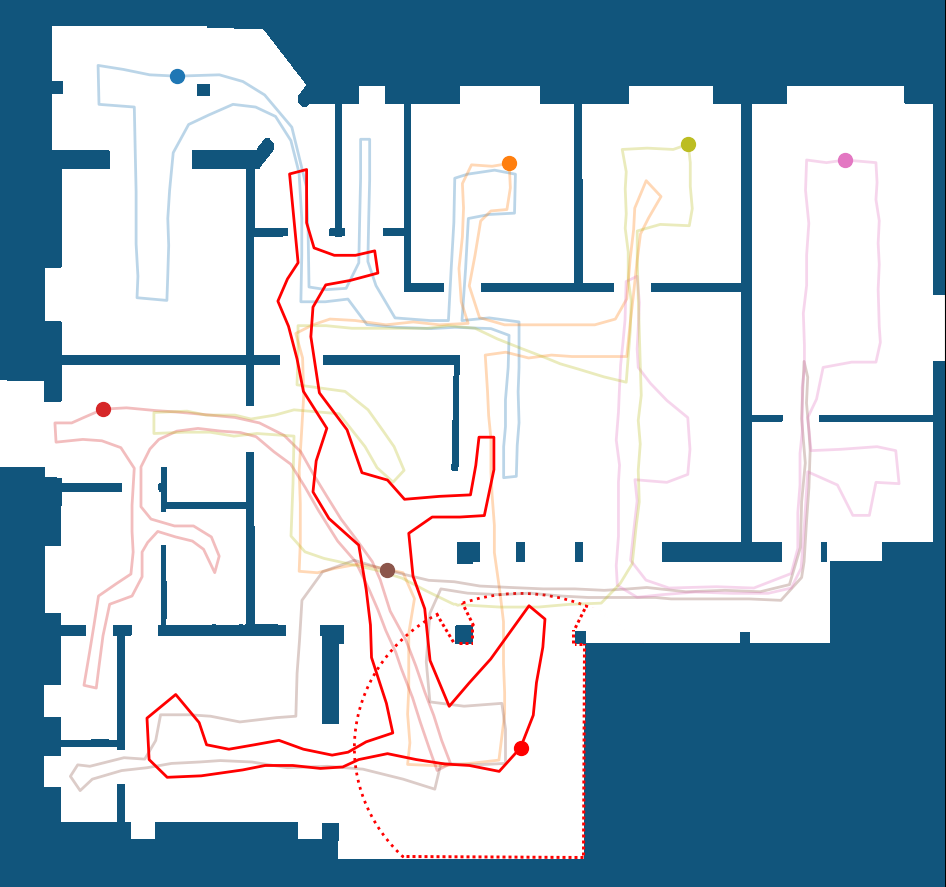}
    \caption{Visualization of the environment map including starting points, exploration pathways, and LiDAR range for the robotic agents as described by \cite{Yu2022}. Adapted from~\cite{Radchenko2024}.}
    \label{fig:env_map}
\end{figure}

In our experiment, we utilized the CubiCasa5K dataset~\cite{Kalervo2019} to generate floor plans for the environment mapping task. Figure~\ref{fig:env_map} illustrates the paths taken by the robots during the mapping process. The solid colored lines represent the trajectories of various robots as they navigate the environment, with the red solid line highlighting a particular robot's path as it explores a designated area. The dotted line surrounding the robot indicates the range of its LiDAR sensor, which acts as a source for the training data.

\section{Peers State Exchange Algorithm}
The original implementation of the DiNNO algorithm was constrained by several limitations that significantly impacted its applicability to edge AI collaborative learning. The original framework featured a sequential approach to the learning process and a centralized experimental architecture, which could benefit from a more distributed design to simulate data exchange better. In this setup, agents were part of a monolithic data structure processed within a single, tightly coupled computational process. This design led to "quasi-agents" that directly accessed each other's memory during the learning process, which, while efficient, did not fully embrace the potential of distributed learning.

To overcome these limitations, we restructured the DiNNO algorithm implementation to make it suitable for edge computing environments. Each agent now operates autonomously, independently processing local LiDAR data and optimizing NN parameters. A key improvement in this updated approach is the introduction of an epoch-based decentralized consensus algorithm, which helps agents exchange NN parameters with each other in a peer-to-peer way (see Algorithm \ref{alg:state_exchange}).    

\begin{algorithm}
\caption{Peers State Exchange~\cite{Radchenko2024}}\label{alg:state_exchange}
\begin{algorithmic}[1]
\Require $MaxRound$, $Socket$, $Id$, $State$
\State \textbf{Initialize:} $Round$, $PeerComplete[ ]$, $PeerState[ ]$
\State $Message \leftarrow (State, 0)$
\State \Call{Send}{$Socket$, $Message$, $Id$}
\While {$Round < MaxRound$}
    \State $(Message, PeerId) \leftarrow$ \Call{Receive}{$Socket$}
    \If {$Message$ is $RoundComplete$}
        \State $PeerComplete[PeerId] \leftarrow \textsc{True}$
    \Else
        \If {$Round < Message.Round$}
            \State \Call{FinishRound}{}
        \EndIf
        \State $PeerState[PeerId] \leftarrow Message.State$
    \EndIf
    \If {$\forall s \in PeerState, s \neq \emptyset$}
        \State $State \leftarrow$ NodeUpdate($State$, $PeerState$)
        \State $\forall s \in PeerState, s \leftarrow \emptyset$
        \State $PeerComplete[Id] \leftarrow \textsc{True}$
        \State $PeerState[Id] \leftarrow State$
        \State $Message \leftarrow RoundComplete$
        \State \Call{Send}{$Socket$, $Message$, $Id$}
    \EndIf
    \If {$\forall p \in PeerComplete, p = \textsc{True}$}
        \State \Call{FinishRound}{}
    \EndIf
\EndWhile
\Function {FinishRound}{}
    \State $\forall p \in PeerComplete, p \leftarrow \textsc{False}$
    \State $Round \leftarrow Round + 1$
    \State $Message.State \leftarrow State$
    \State $Message.Round \leftarrow Round$
    \State \Call{Send}{$Socket$, $Message$, $Id$}
\EndFunction
\end{algorithmic}
\end{algorithm}

The peers state exchange algorithm starts with the following inputs: the maximum number of synchronization epochs ($MaxRound$), network socket ($Socket$), unique peer identifier ($Id$), and the initial state of the NN parameters ($State$). The algorithm uses two data structures to keep track of the communications process: $PeerComplete[]$, which monitors whether each peer has finished a round, and $PeerState[]$, which stores the current NN parameters for each peer.

The algorithm revolves around the P2P exchange of two message types: $State$ and $RoundComplete$. The $RoundComplete$ message signals the completion of a round by a peer, while the $State$ message contains the peer’s NN parameters state for the current round. Including the $RoundComplete$ message along with a round finalization logic addresses the challenges posed by network latency — such as out-of-order messages, delayed status updates, and potential desynchronization between peers. To mitigate the latency issues, the algorithm ensures that a $RoundComplete$ message is dispatched by a peer only after successfully receiving all $State$ messages from the other peers. This ensures that no peer progresses to the subsequent round until all peers have synchronized on the current round, preventing the risk of desynchronization caused by delayed or missing messages. 

When a peer receives a $State$ message from a future round, the algorithm triggers the $\textsc{FinishRound}$ function, prompting the peer to align with the correct round. This mechanism manages out-of-order deliveries, ensuring consistency and synchronization across all agents.

This version of the algorithm assumes that all messages will eventually reach their intended recipients, excluding the consideration of scenarios involving agent malfunctions, computational halts, or permanent network failures that could result in irreversible message loss or total communication breakdown.

\section{Distributed Uncertainty Estimation}
\begin{figure}[htbp]
    \centering
    \includegraphics[width=0.3\textwidth]{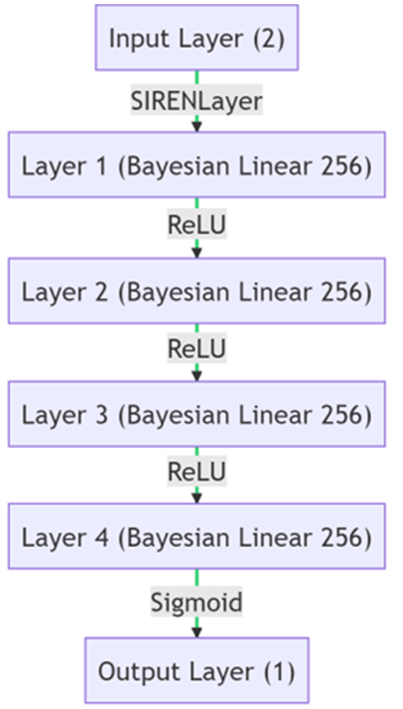}
    \caption{Visualization of the proposed collaborative mapping BNN architecture.}
    \label{fig:bnn}
\end{figure}

To address uncertainty estimation in the distributed mapping problem, we developed a BNN incorporating Bayesian Linear Layers in the neural network (see Figure~\ref{fig:bnn}). 

The architecture of the BNN is detailed as follows:
\begin{itemize}
    \item \textbf{Input Layer (2):} The input consists of coordinates $x, y$, representing a global position on the map.
    \item \textbf{SIREN Layer (256):} A layer with a sinusoidal activation function, designed for Neural Implicit Mapping, as described by~\cite{Sitzmann2020}.
    \item \textbf{4 x Bayesian Linear Layers (256):} Four Bayesian Linear Layers, each with 256 nodes and activated by the ReLU function. These layers are probabilistic and support uncertainty estimation.
    \item \textbf{Output Layer (1):} A linear layer with one node, activated by the Sigmoid function. An output of $0$ indicates empty space, while an output of $1$ indicates occupied space (e.g., a wall).
\end{itemize}

Unlike deterministic networks that produce only a single output, BNNs are distinguished by their ability to estimate prediction uncertainty. By conducting multiple forward passes to compute the mean and standard deviation of the outputs, BNNs provide valuable insights into the model's confidence for each mesh grid point. To correctly regularize the BNN parameters during the distributed learning phase, we developed Algorithm~\ref{alg:bnns_opt}, which takes into account the specific roles of the median ($\mu$) and standard deviation ($\rho$) parameters of BNN neurons. For regularizing the BNN $\rho$-parameters across models of individual actors, we employ KL Divergence, as detailed in Equation (6).

\begin{algorithm}
\caption{Optimization of BNN Parameters~\cite{Radchenko2024}}\label{alg:bnns_opt}
\begin{algorithmic}[1]
\Require $Model$, $Optimizer_{\mu}$, $Optimizer_{\rho}$, $W_{\mu}$, $W_{\rho}$, $Iter$, $\theta_{\text{reg}\mu}$, $\theta_{\text{reg}\rho}$, $Duals_{\mu}$, $Duals_{\rho}$
\For{$i \leftarrow 1$ to $Iter$}
    \State Reset gradients of $Optimizer_{\mu}$ and $Optimizer_{\rho}$
    \State $PredLoss \leftarrow$ \textsc{ComputeLoss}($Model$)
    \State $\theta_{\mu}, \theta_{\rho} \leftarrow$ \textsc{ExtractParameters}($Model$)
    \State $Reg_{\mu} \leftarrow$ \textsc{L2Regularization}($\theta_{\mu}, \theta_{\text{reg}\mu}$)
    \State $Reg_{\rho} \leftarrow D_{KL}(\theta_{\rho}, \theta_{\text{reg}\rho})$
    \State $Loss_{\mu} \leftarrow PredLoss + \langle \theta_{\mu}, Duals_{\mu} \rangle + W_{\mu} \times Reg_{\mu}$
    \State $Loss_{\rho} \leftarrow \langle \theta_{\rho}, Duals_{\rho} \rangle + W_{\rho} \times Reg_{\rho}$
    \State \textsc{UpdateParameters}($Optimizer_{\mu}$, $Loss_{\mu}$)
    \State \textsc{UpdateParameters}($Optimizer_{\rho}$, $Loss_{\rho}$)
\EndFor
\end{algorithmic}
\end{algorithm}

\section{Implementation and Evaluation of the Edge AI Collaborative Learning}
\begin{figure}[htbp]
    \centering
    \includegraphics[width=0.6\textwidth]{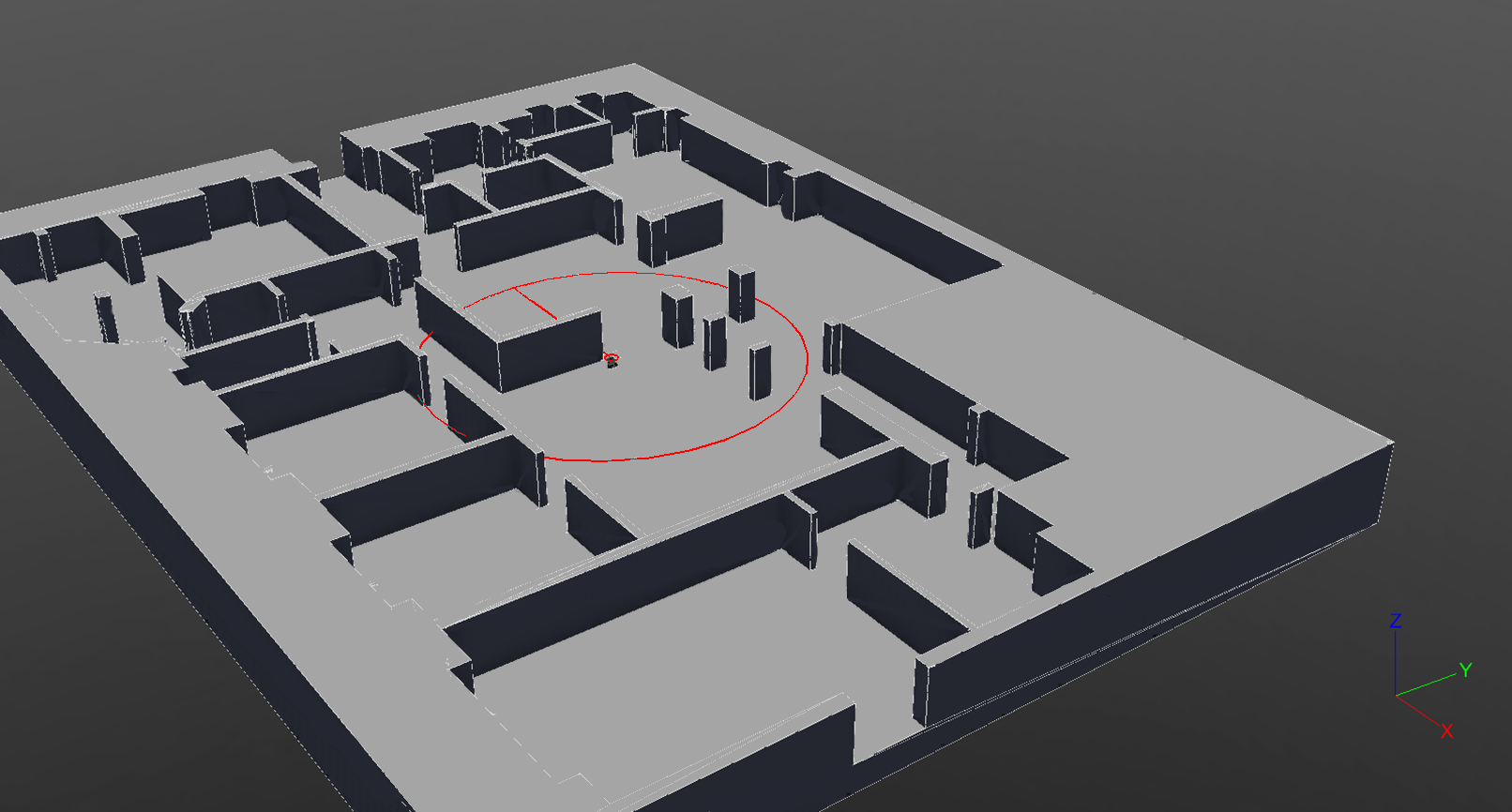}
    \caption{Visualization of the 3D model of the environment generated from the floor plan showcasing a LiDAR-equipped TurtleBot navigating the space in a Webots simulation~\cite{Radchenko2024}.}
    \label{fig:robot_nav}
\end{figure}

To simulate a realistic environment for robotic exploration, we utilized floor plan data from the CubiCasa5K dataset to generate three-dimensional interior models in STL format. These models were subsequently imported into the Webots simulation platform, where TurtleBot robots were programmed to navigate through the space (see Figure~\ref{fig:robot_nav}). In our simulation, the TurtleBots were equipped with essential sensors, including LiDAR, which introduced realistic sensor noise and measurement uncertainties.

The environment was designed to represent real-world conditions, allowing for the testing of mapping algorithms. The experiment involved launching seven independent TurtleBot agents that gradually collected information from LiDAR sensors while exploring a virtual interior space. TurtleBot movement was controlled by a Python-based script that guided the robots through the environment. This controller managed robot orientation and movement using a compass and GPS data. The study assumed that all robots had access to global positioning information.

Agents' movement paths were predefined to create simulation programs for their interior navigation. LiDAR sensor data collection was simulated as a Webots data stream during navigation. The LiDAR sensor scanned the surroundings and updated an occupancy grid map, where each cell was classified as unknown, free, or occupied. LiDAR data was transformed to fit the Neural Implicit Mapping framework, converting raw scans into a point set with values ranging from 0 to 1. In this mapping, a value of 1 indicated the presence of a wall, while 0 represented empty space.

Within the experiment's framework, each agent ran as a separate Python process. Agent communication was implemented via direct TCP connections between processes on the same virtual local network. The ZeroMQ library was used for asynchronous data exchange. Containerization of agent processes was achieved using Singularity containers provided with GPU access. In the experiments outlined, all processes were initiated on GPU-enabled computing nodes managed by the SLURM workload manager.

\subsection{Uncertainty Evaluation: Individual Agent Case}
\label{sec:un_individual}
\begin{figure}[htbp]
    \centering
    \begin{subfigure}[t]{0.3\textwidth}
        \includegraphics[width=\textwidth]{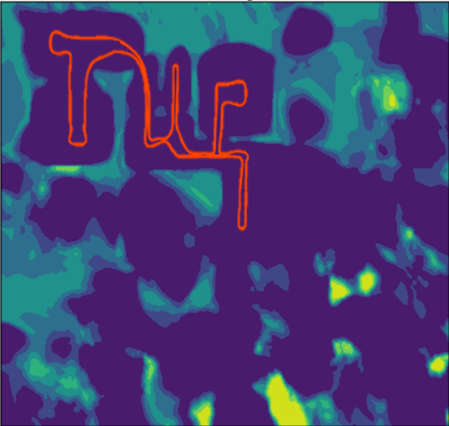}
        \caption{Single forward pass}
        \label{subfig:single_forward}
    \end{subfigure}
    \begin{subfigure}[t]{0.3\textwidth}
        \includegraphics[width=\textwidth]{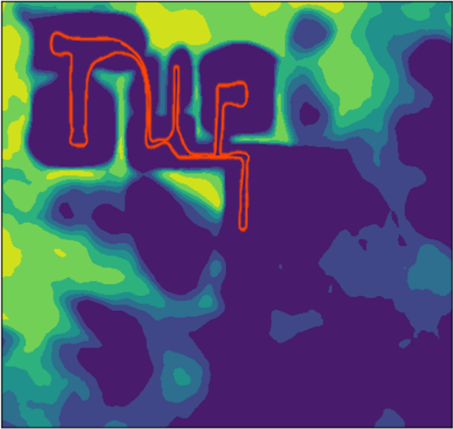}
        \caption{Mean of 50 forward passes}
        \label{subfig:single_mean_50_forward}
    \end{subfigure}
    \begin{subfigure}[t]{0.3\textwidth}
        \includegraphics[width=\textwidth]{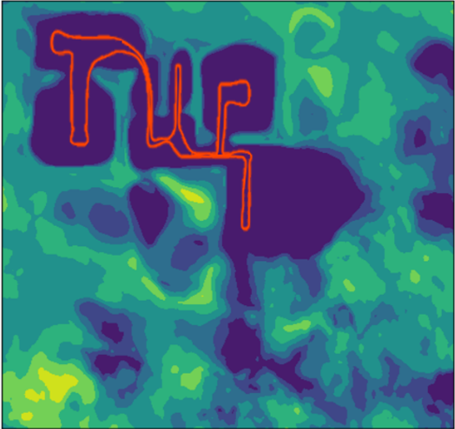}
        \caption{Standard deviations of 50 forward passes}
        \label{subfig:single_dev_50_forward}
    \end{subfigure}
    \caption{Visualization of the individual agent BNN training results. The red line represents the path of the single agent through the environment.}
    \label{fig:single_bnn}
\end{figure}

To evaluate the effectiveness of the BNN architecture proposed in Section 5 for estimating uncertainty in NN outcomes, we designed an experiment to demonstrate how the BNN captures and represents uncertainty in its predictions, particularly in areas where the agent has no prior data. The robotic agent was trained exclusively on local data during the experiment without exchanging information with other agents. The visualization of the training results is presented in Figure~\ref{fig:single_bnn}. 

To generate outputs from the BNN, 50 queries were made for each pair of input coordinates $(x, y)$. Subsequently, a visualization was created to illustrate the mean values and standard deviations of the NN responses. The red line on the graph indicates the actual path taken by the agent, while the surrounding colors represent the network's prediction of the environment's state based on the agent's current understanding. 

Subplot~~\ref{subfig:single_forward} depicts the results of a single forward pass through the BNN. This visualization displays a significant amount of noise in regions the agent has not explored, indicating high uncertainty in those areas. Subplot~\ref{subfig:single_mean_50_forward} shows the mean of 50 forward passes, resulting in a much smoother and more accurate representation of the investigated environment, highlighting the BNN's capability to refine its predictions by reducing uncertainty through repeated passes. Finally, subplot~\ref{subfig:single_dev_50_forward} illustrates the standard deviation across the 50 forward passes, effectively visualizing the uncertainty associated with the network's predictions. Higher standard deviations indicate regions where the network is less certain about its predictions, often corresponding to areas with less data or where the agent has less experience. 

Following this initial test, we conducted a series of experiments to evaluate the impact of the $kl_{weight}$ parameter from Equation (7) on the training results. The visualization of the experiments results is presented in Figure~\ref{fig:uncertainty_estimation}.

\begin{figure}[htbp]
    \centering
    \begin{subfigure}[t]{0.3\textwidth}
        \includegraphics[width=\textwidth]{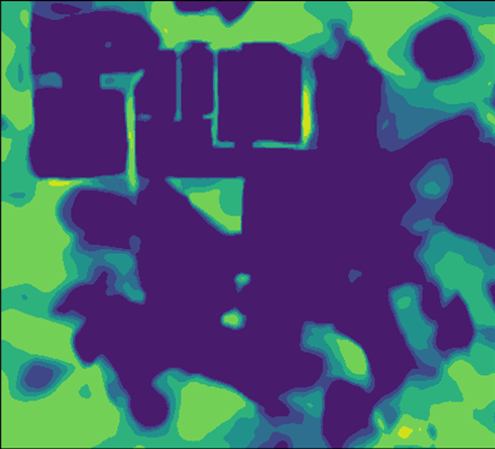}
        \caption{Mean $kl_{weight}=10^{-4}$}
    \end{subfigure}
    \begin{subfigure}[t]{0.29\textwidth}
        \includegraphics[width=\textwidth]{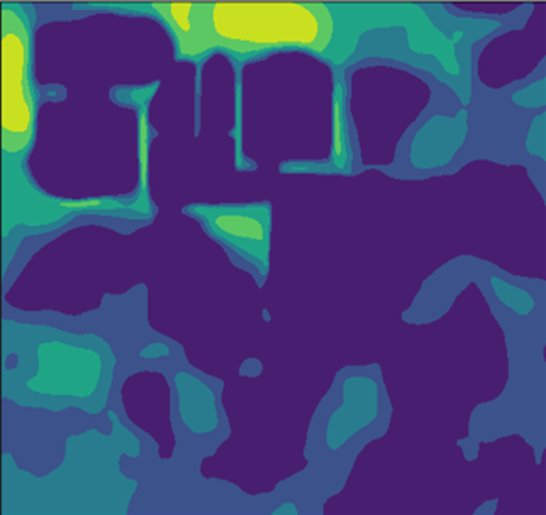}
        \caption{Mean $kl_{weight}=5 \times 10^{-3}$}
    \end{subfigure}
    \begin{subfigure}[t]{0.298\textwidth}
        \includegraphics[width=\textwidth]{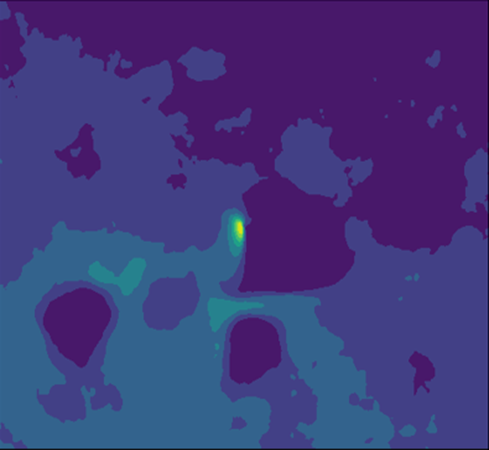}
        \caption{Mean $kl_{weight}=5 \times 10^{-1}$}
    \end{subfigure}
    
    \begin{subfigure}[t]{0.3\textwidth}
        \includegraphics[width=\textwidth]{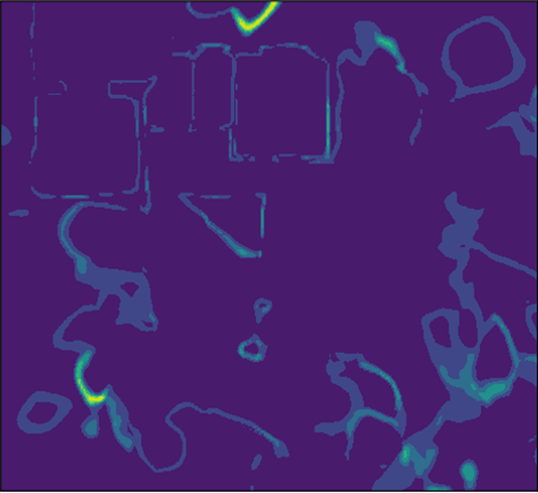}
        \caption{Standard deviation $kl_{weight}=10^{-4}$}
    \end{subfigure}
    \begin{subfigure}[t]{0.295\textwidth}
        \includegraphics[width=\textwidth]{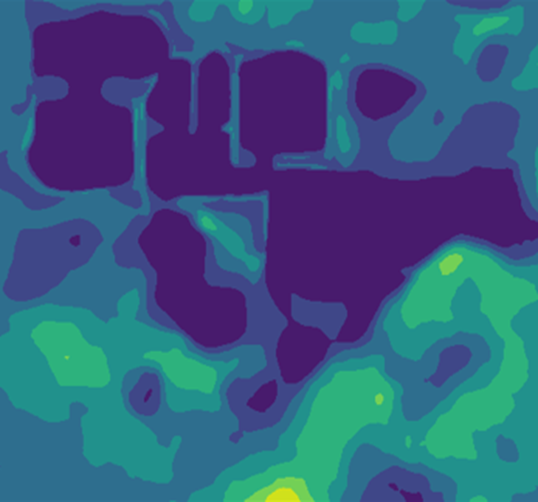}
        \caption{Standard deviation $kl_{weight}=5 \times 10^{-3}$}
    \end{subfigure}
    \begin{subfigure}[t]{0.305\textwidth}
        \includegraphics[width=\textwidth]{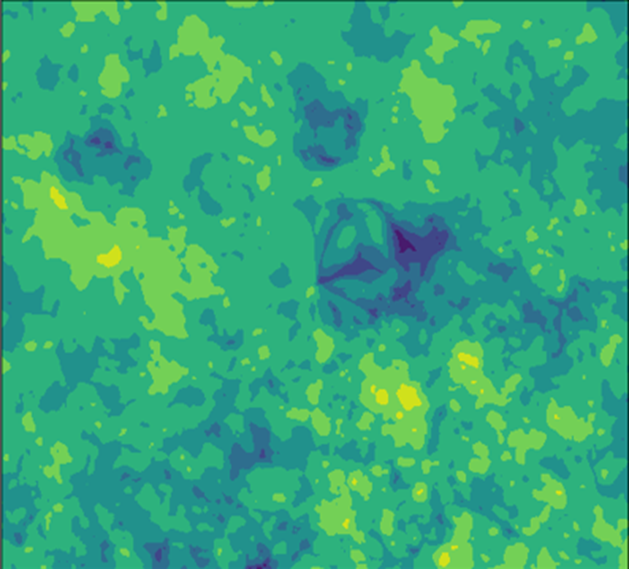}
        \caption{Standard deviation $kl_{weight}=5 \times 10^{-1}$}
    \end{subfigure}
    \caption{Comparative visualization of the influence of the $kl_{weight}$ parameter on the single-agent uncertainty estimation. Adapted from~\cite{Radchenko2024}.}
    \label{fig:uncertainty_estimation}
\end{figure}

A low value of the $kl_{weight}$ parameter was observed to lead to a low variance in the NN’s results, which does not allow for distinguishing the hallucinations of the NN from areas with sufficient data to form a general understanding of the environment.  onversely, a high $kl_{weight}$ parameter value results in excessive noise and high uncertainty in the NN’s results. Therefore, to ensure that the BNN provides an effective assessment of uncertainty, fine-tuning the $kl_{weight}$ parameter during the training process is required.

\subsection{Online Learning Process Evaluation}

Previously, we explored the results derived from analyzing data sets collected and processed collectively after the agents completed their traversal. This approach is necessary due to the substantial processing power and energy resources required for NN training, which may not be readily available to autonomous edge devices operating in a mobile investigation mode.

To evaluate the applicability of online training — combining agent exploration with the learning process — we implemented a simulation that used real-time data gathering alongside a batch training approach. This simulation was built using the ZeroMQ library to implement the communication between components. A streamer broker was initiated as a subprocess from the Webots controller, while the NN training process was launched as a parallel subprocess, acting as the data receiver. The data-sending function was embedded directly within the Webots controller. As the robot traversed its path, it periodically sent LiDAR data to the broker after scanning a set number of positions. The NN training process then added this data to the training pool, allowing for a gradual compilation of training data as the path was navigated.

The experiment aimed to assess the effectiveness of the training process based on partial data collection. To structure this process, we introduced the "Communication Rounds" (CR) concept, which segmented the path into discrete parts, each representing a phase of data collection and subsequent training. The agent gathered LiDAR data for each CR segment and transmitted this information to the training algorithm for partial data training at the segment's end.

We specifically investigated two scenarios in our experiments:
\begin{itemize}
    \item \textbf{Cumulative Data Retention (CDR)}: In this scenario, the algorithm retained all previously received data, using it alongside new data for subsequent training sessions. This method was intended to continuously enrich the training dataset, potentially enhancing the NN's accuracy and adaptability.
    \item \textbf{Data Refresh (DR)}: In contrast, this approach trained the NN solely on the newly acquired dataset for each CR, discarding all previous datasets. The only remnants of prior data were the pre-trained weights of the NN, which could influence the training outcome based on past learning but did not directly reuse old data.
\end{itemize}

\begin{figure}[t]
    \centering
    \begin{subfigure}[t]{0.24\textwidth}
        \centering
        \includegraphics[width=\textwidth]{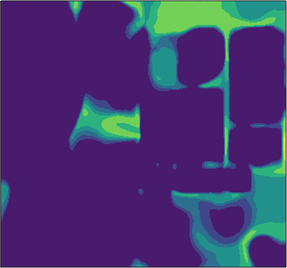}
    \end{subfigure}
    \hfill
    \begin{subfigure}[t]{0.24\textwidth}
        \centering
        \includegraphics[width=\textwidth]{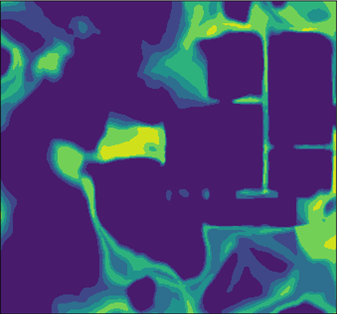}
    \end{subfigure}
    \hfill
    \begin{subfigure}[t]{0.24\textwidth}
        \centering
        \includegraphics[width=\textwidth]{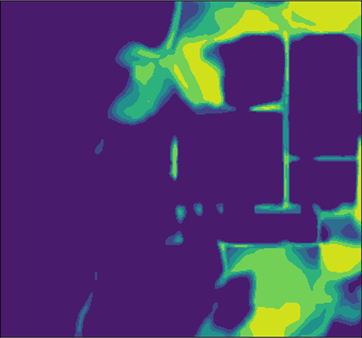}
    \end{subfigure}
    \hfill
    \begin{subfigure}[t]{0.24\textwidth}
        \centering
        \includegraphics[width=\textwidth]{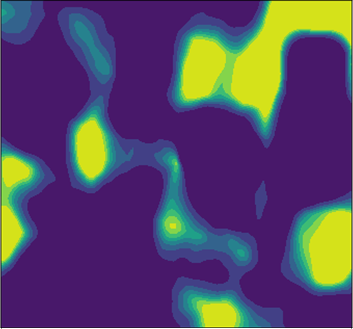}
    \end{subfigure}
    
    \vspace{0.5cm}
    \begin{subfigure}[t]{0.24\textwidth}
        \centering
        \includegraphics[width=\textwidth]{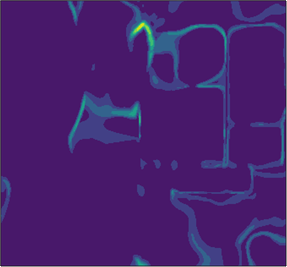}
        \caption{CDR, 8 CR}
        \label{fig:online_8_std}
    \end{subfigure}
    \hfill
    \begin{subfigure}[t]{0.24\textwidth}
        \centering
        \includegraphics[width=\textwidth]{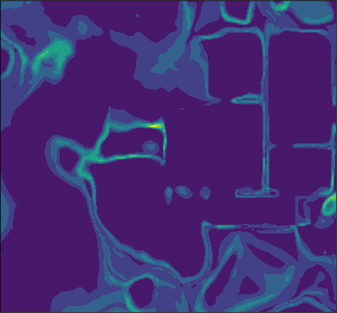}
        \caption{CDR, 18 CR}
        \label{fig:online_18_std}
    \end{subfigure}
    \hfill
    \begin{subfigure}[t]{0.24\textwidth}
        \centering
        \includegraphics[width=\textwidth]{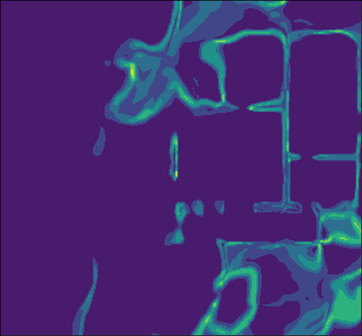}
        \caption{CDR, 25 CR}
        \label{fig:online_25_std}
    \end{subfigure}
    \hfill
    \begin{subfigure}[t]{0.24\textwidth}
        \centering
        \includegraphics[width=\textwidth]{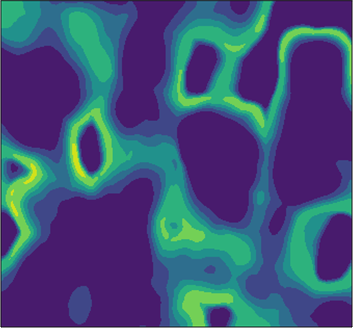}
        \caption{DR, 18 CR}
        \label{fig:online_refresh_18_std}
    \end{subfigure}
    
    \caption{Visualization of the Mean (top row) and Standard Deviation (bottom row) of the online training process results under different configurations of Cumulative Data Retention (CDR) and Data Refresh (DR) scenarios.}
    \label{fig:online_training_results}
\end{figure}

The results of these experiments are visualized in Figure~\ref{fig:online_training_results}. Each column in the figure corresponds to different configurations of the online training process, with varying numbers of communication rounds and the implementation of either Cumulative Data Retention (CDR) or Data Refresh (DR) strategies. As the number of communication rounds increases, the quality of the environment mapping improves, reflecting more accurate mean predictions and reduced standard deviations. In the CDR scenario, the NN demonstrates better stability and consistency in its predictions across all communication rounds, indicating that the accumulation of data significantly benefits the learning process. On the other hand, in the DR scenario, while initial rounds show promising results, later rounds exhibit increased noise and uncertainty, suggesting that the DR strategy may lead to a significant loss of valuable information, affecting the overall performance of the NN.

\subsection{Uncertainty Evaluation: Multi-Agent Case}
In section~\ref{sec:un_individual}, we demonstrated that BNNs can effectively estimate uncertainty in the context of a single learning agent. In this section, we extend that analysis to explore whether the same capabilities of BNNs are in decentralized multi-agent training environments. To investigate this, we evaluated different regularization strategies to understand their impact on the quality of decentralized BNN training, with a particular focus on assessing validation loss under the following approaches: 

\begin{enumerate}
    \item Applying a uniform L2 regularization across all neural network parameters without differentiating between parameter types;
    \item Applying separate L2 regularization for standard neural network parameters and Bayesian parameters;
    \item Using distinct regularization techniques for standard and Bayesian parameters, with L2 regularization for the standard parameters and Kullback-Leibler divergence for the Bayesian parameters (see Algorithm~\ref{alg:bnns_opt}).
\end{enumerate}

\begin{figure}[htbp]
    \centering
    \includegraphics[width=0.7\textwidth]{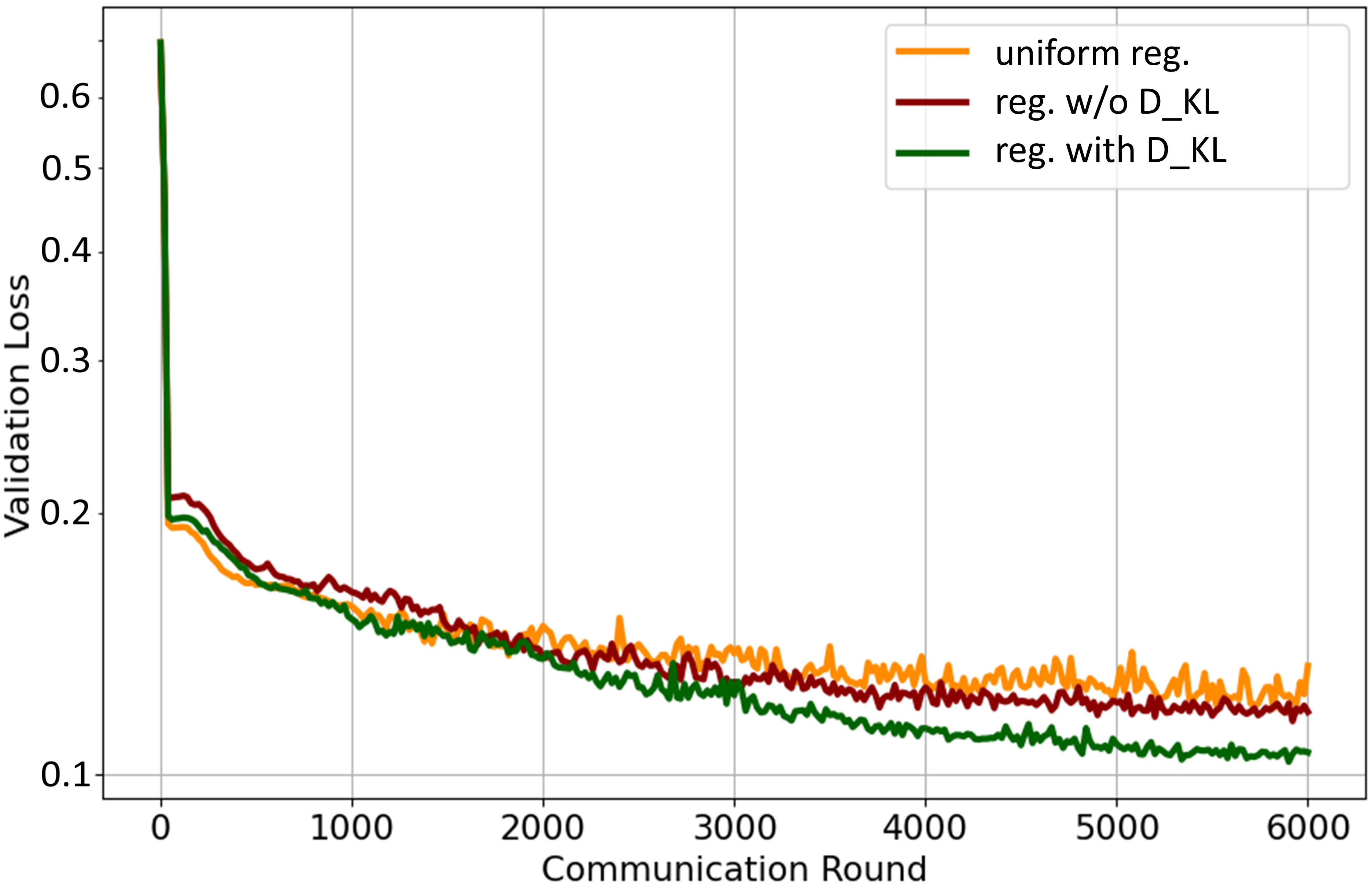}
    \caption{Comparison of validation loss during distributed BNN training 1) with uniform L2 regularization (\textbf{uniform reg.}); 2) separate L2 regularization (\textbf{reg. w/o D\_KL}); 3) Kullback-Leibler divergence for regularization of BNN $\rho$-parameters (\textbf{reg. with D\_KL})~\cite{Radchenko2024}.}
    \label{fig:validation_loss}
\end{figure}

The findings from these evaluations are illustrated in Figure~\ref{fig:validation_loss}. Notably, the use of KL divergence for regularizing Bayesian parameters, as detailed in Algorithm~\ref{alg:bnns_opt}) led to a 12-30\% reduction in validation loss compared to the other approaches. This strategy also enhanced the stability of the training process. 

\begin{figure}[htbp]
    \centering
    \begin{subfigure}[b]{0.295\textwidth}
        \includegraphics[width=\textwidth]{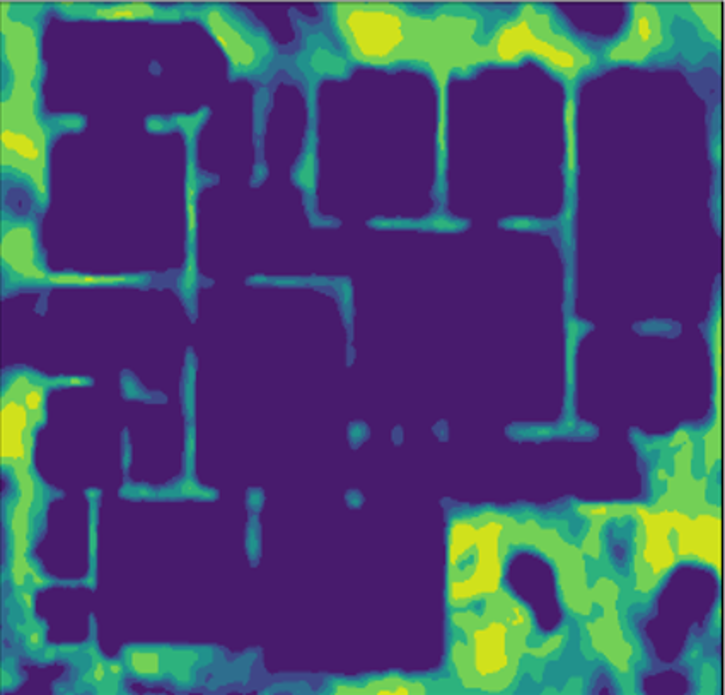}
        \caption{Mean}
        \label{subfig:dec_mean}
    \end{subfigure}
    \begin{subfigure}[b]{0.3\textwidth}
        \includegraphics[width=\textwidth]{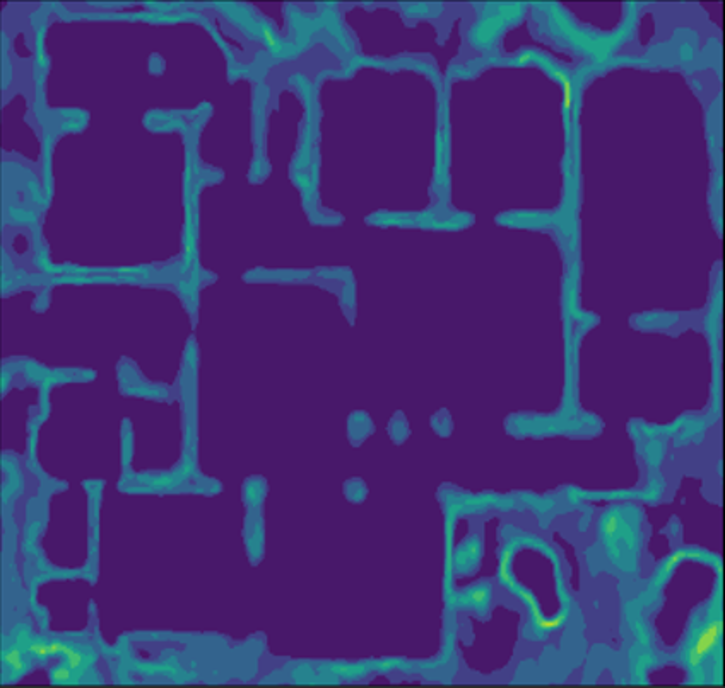}
        \caption{Standard deviation}
        \label{subfig:dec_dev}
    \end{subfigure}
    \begin{subfigure}[b]{0.285\textwidth}
        \includegraphics[width=\textwidth]{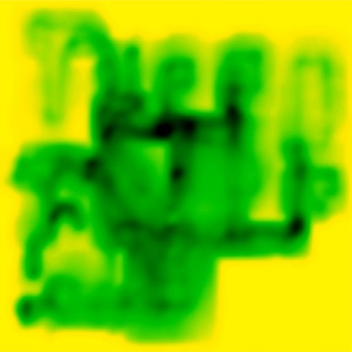}
        \caption{Point density}
        \label{subfig:dec_point}
    \end{subfigure}
    \caption{Visualization of the decentralized BNN training results according to Algorithm \ref{alg:bnns_opt}: (a) mean; (b) standard deviation; (c) point density of the original data set. Adapted from~\cite{Radchenko2024}.}
    \label{fig:decentralized_bnn}
\end{figure}

The results of the decentralized BNN training using this method are visualized in Figure~\ref{fig:decentralized_bnn}.

The figure depicts the results of 50 forward passes of a generalized model that the seven individual agents converged to during the training process utilizing Algorithms~\ref{alg:state_exchange} and~\ref{alg:bnns_opt}. During the training, all agents reached a consensus on a unified model. Subfigure~\ref{subfig:dec_mean} shows the mean predictions across the environment, representing a general map that encapsulates the knowledge gathered by all seven agents. Subfigure~\ref{subfig:dec_dev} illustrates the standard deviation of these predictions, which can be interpreted as the level of uncertainty associated with the model's output. Higher standard deviations indicate regions where the model is less certain about its predictions. 

As an uncertainty baseline, subfigure~\ref{subfig:dec_point} shows the density of LiDAR data points used for model training. In this subfigure, darker areas represent regions with a higher concentration of data points, offering a reference for comparing the uncertainty visualized in subfiguree~\ref{subfig:dec_dev} against the actual data distribution. To estimate the density of the LiDAR data points, we employed the Kernel Density Estimation method~\cite{Scott_D2015} with a Gaussian kernel.

\section{Conclusions}
This paper addressed the problem of uncertainty estimation in edge collaborative learning. Through a targeted case study on collaborative mapping, we developed and evaluated a decentralized learning framework based on the Webots simulation platform and Distributed Neural Network Optimization algorithm. Our main contributions include the design of an epoch-based decentralized consensus algorithm to support the peer-to-peer exchange of neural network parameters among distributed agents and integrating BNNs to incorporate uncertainty estimation into this decentralized learning context.

We reviewed EL methods and their applicability to AI-enabled edge devices alongside our practical implementations. This review covered several approaches, including Federated Learning, Federated Distillation, Split Learning, and decentralized techniques based on the Alternating Direction Method of Multipliers. We examined the benefits and drawbacks of these methods and discussed how they could be adapted and applied to enhance edge learning environments. Additionally, we explored how BNNs could be used for uncertainty estimation, emphasizing the potential of stochastic modeling to improve the generalization capabilities of neural networks in complex and uncertain scenarios.

Our findings show that BNNs are effective in estimating uncertainty within distributed learning environments, but they also highlight the importance of precise hyperparameter tuning to achieve reliable uncertainty assessments. Among the regularization strategies evaluated, using Kullback–Leibler divergence for Bayesian parameter regularization was particularly beneficial, resulting in a 12-30\% reduction in validation loss compared to other strategies. This approach also contributed to greater stability during training, demonstrating the practical value of integrating BNNs into distributed learning frameworks.

Future work should focus on optimizing these distributed learning techniques for implementation on embedded AI hardware, where computational resources are limited. This will require refining neural network architectures and EL methods to fit the specific constraints of edge devices. Furthermore, there is a need to explore advanced task management and offloading strategies within the multi-layered fog and hybrid edge-fog-cloud infrastructures to enhance computational efficiency and resource utilization in dynamic and resource-constrained environments.

\bibliographystyle{unsrt}  
\bibliography{closer}

\end{document}